\documentclass[preprint,12pt]{elsarticle}

\usepackage{amssymb}
\usepackage{amsmath}

\journal{}

\begin{document}

\begin{frontmatter}



\title{Unsupervised categorization of similarity measures}


\author[af1]{Yoshiyuki Ohmura\corref{cor1}} 
\ead{ohmura@isi.imi.i.u-tokyo.ac.jp}
\cortext[cor1]{Yoshiyuki Ohmura}
\author[af1]{Wataru Shimaya}
\author[af1]{Yasuo Kuniyoshi}

\affiliation[af1]{organization={Department of Mechano-Informatics, Graduate School of Information Science and Technology, The University of Tokyo, },
            addressline={7-3-1, Hongo, Bunkyo-ku}, 
            city={Tokyo},
            postcode={113-8656}, 
            country={Japan}}

\begin{abstract}
In general, objects can be distinguished on the basis of their features, such as color or shape.
In particular, it is assumed that similarity judgments about such features can be processed independently in different metric spaces. 
However, the unsupervised categorization mechanism of metric spaces corresponding to object features remains unknown. 
Here, we show that the artificial neural network system can autonomously categorize metric spaces through representation learning to satisfy the algebraic independence between neural networks, and project sensory information onto multiple high-dimensional metric spaces to independently evaluate the differences and similarities between features. 
Conventional methods often constrain the axes of the latent space to be mutually independent or orthogonal.  
However, the independent axes are not suitable for categorizing metric spaces. 
High-dimensional metric spaces that are independent of each other are not uniquely determined by the mutually independent axes, because any combination of independent axes can form mutually independent spaces.
In other words, the mutually independent axes cannot be used to naturally categorize different feature spaces, such as color space and shape space. 
Therefore, constraining the axes to be mutually independent  makes it difficult to categorize high-dimensional metric spaces.
To overcome this problem, we developed a method to constrain only the spaces to be mutually independent and not the composed axes to be independent. Our theory provides general conditions for the unsupervised categorization of independent metric spaces, thus advancing the mathematical theory of functional differentiation of neural networks. 
\end{abstract}



\begin{keyword}
 Algebraic independence \sep invariant transformation \sep metric space categorization
\end{keyword}

\end{frontmatter}



\section{Introduction}
\label{sec1}
The similarity between objects is determined based on their features, such as color, shape, and size.
Judging the similarity between different types of features is difficult for humans, but features of the same type are easy to compare. 
For example, we can see that red is more similar to pink than to black. 
However, we cannot judge whether the red is more similar to a triangle than to a square. 
These similarity judgments are considered to be processed independently in different areas of the brain\cite{Taylor2022}. 
How does the functional differentiation for similarity judgments develop in the neural network system?

Unsupervised clustering methods can be used to categorize objects based on a single predefined similarity measure\cite{Jain2010,Taha2023}, but not to categorize the similarity measures themselves. 
How the neural network system categorizes similarity measures corresponding to different sensory features, such as color or shape, remains unknown. 
Recently, Fumero et al. provided a method for categorizing feature similarities using weakly supervised learning, rather than unsupervised learning\cite{Fumero2021}. 
They assumed that there are multiple transformations corresponding to each feature between visual scenes, and that the dataset used consists of pairs of visual scenes whose difference can be represented by only one of the features. 
Their method is valid only when the dataset satisfies this assumption and is not applicable when there are multiple feature differences between pairs of scenes.
Thus, the unsupervised categorization method of similarity measures remains unknown. 

Here, we show that the neural network system can autonomously categorize the metric spaces through representation learning to satisfy the algebraic independence\cite{Simpson2018} between neural networks. Algebraic independence is a generalized form of independence that includes stochastic independence and orthogonality. 
Conventional representation learning methods often decompose sensory input into mutually independent axes \cite{higgins2017}. However, the independent axes are not suitable for forming mutually independent spaces, such as a three-dimensional color space and a high-dimensional shape space. 
If all axes are mutually independent, then any two spaces composed of the independent axes are mutually independent, suggesting that independent axes cannot be used to define unique mutually independent spaces. 
In order to define mutually independent feature spaces, we developed a new method to constrain only the space to be independent without constraining the composed axes to be independent. Our contribution is to propose the effectiveness of constraining the algebraic structure among neural networks rather than the distribution of data.

\section{Formulation: A relation between the algebraic independence and the invariant transformation equation}
The algebraic independence structure in category theory\cite{Simpson2018} is a generalization of several independence models in mathematics, including the stochastic independence which is often used in representation learning\cite{higgins2017,Hyvarinen2000,higgins2018}. 
Here, based on this algebraic independence in category theory\cite{Simpson2018}, we develop an algebraic independence between transformations and show a connection to the invariant transformation equation, which is often used to formulate pattern recognition theory\cite{Otsu1986, takada2021} and representation theory\cite{higgins2018}. 

Invariant transformation means that if two feature vectors, $\boldsymbol{\lambda}_0$ and $\boldsymbol{\lambda}_1$, are independent of each other, then a $\boldsymbol{\lambda}_0$-transformation changes only $\boldsymbol{\lambda}_0$, keeping with $\boldsymbol{\lambda}_1$ invariant. 
Similarly, a $\boldsymbol{\lambda}_1$-transformation changes only $\boldsymbol{\lambda}_1$. 
The invariant transformation equation is associated with perceptual constancy and is commonly used in the formulation of pattern recognition. 

In the following formulation, we limit the number of transformations to two for simplicity; however, a formulation allowing an arbitrary number of transformations can be obtained from the present formulation (Supplementary Information).

We define the observation space as a countable set, $S$ of $\mathbb{R}^N$. Let the relationship between two points, $X,Y \in S$, be denoted as $Y = F_0 (\boldsymbol{\lambda}_0 ) F_1 (\boldsymbol{\lambda}_1 )[X] $, where $F_0$ and $F_1$ are transformations (functions whose input and output dimensions are the same) and $\boldsymbol{\lambda}_0 \in \mathbb{R}^{n_0} $ and $\boldsymbol{\lambda}_1 \in \mathbb{R}^{n_1}$ are transformation parameters. If the two transformations, $F_0 (\boldsymbol{\lambda}_0)$ and $F_1 (\boldsymbol{\lambda}_1)$, satisfy the following conditions, they are algebraically independent.

\begin{itemize}
\item Condition 1 (the existence of unit element): 

$\exists \boldsymbol{\lambda}_0^I:F_0 (\boldsymbol{\lambda}_0^I )=I$ and $\exists \boldsymbol{\lambda}_1^I:F_1 (\boldsymbol{\lambda}_1^I )=I$, where $I$ is the identity transformation. Hereinafter, the unit transformation parameters are denoted as 0. From condition 1, we have an invariant condition of the transformation, such as $\forall \boldsymbol{\lambda}_0:F_0 (\boldsymbol{\lambda}_0) F_0 (0)=F_0 (\boldsymbol{\lambda}_0 )$.
\item Condition 2 (commutativity): 

$F_0 (\boldsymbol{\lambda}_0 ) F_1 (\boldsymbol{\lambda}_1 )=F_1 (\boldsymbol{\lambda}_1 ) F_0 (\boldsymbol{\lambda}_0 )$. From condition 2, the unique composition of two transformations, $Y$, can be obtained from $X$ and transformation parameters, $\boldsymbol{\lambda}_0$ and $\boldsymbol{\lambda}_1$. 
\item Condition 3 (the uniqueness of transformation parameters):

$Y=F_0 (\boldsymbol{\lambda}_0 ) F_1 (\boldsymbol{\lambda}_1 )[X]=F_1 (\boldsymbol{\lambda}_1 ) F_0 (\boldsymbol{\lambda}_0 )[X]=\Phi (\boldsymbol{\lambda}_0,\boldsymbol{\lambda}_1,X)$, where $\Phi$ is a bijective function. In other words, the transformation parameters, $\boldsymbol{\lambda}_0$ and $\boldsymbol{\lambda}_1$, are uniquely determined from $X$ and $Y$. 
\end{itemize}

When we change the transformation parameter, $\boldsymbol{\lambda}_0$, to $\boldsymbol{\lambda}_0^{'}$, we uniquely obtain $Y^{'}=\Phi(\boldsymbol{\lambda}_0^{'},\boldsymbol{\lambda}_1,X)$. Therefore, the transformation F$_0$ is a $\boldsymbol{\lambda}_1$-invariant transformation. Similarly, the transformation $F_1$ is a $\boldsymbol{\lambda}_0$-invariant transformation. Thus, the algebraic independence between transformations defines the invariant transformation equation.

\section{Categorization of metric spaces based on algebraic independence}
We construct multiple metric spaces to measure similarity using two norms of the transformation parameters, $\| \boldsymbol{\lambda}_0 \| $ and $\| \boldsymbol{\lambda}_1 \|$. Therefore, the points $X$ and $Y$ project to two latent spaces, $Q_0 \in \mathbb{R}^{n_0}$ and $Q_1 \in \mathbb{R}^{n_1 }$, respectively. Then, we define the transformation parameters, $\boldsymbol{\lambda}_0$ and $\boldsymbol{\lambda}_1$, by subtraction of latent vectors corresponding to $X$ and $Y$. We employed a familiar encoder-decoder model in representation learning\cite{Fumero2021, higgins2017}. Both encoder and decoder are frequently implemented using neural networks. Theorem 1 provides the conditions to satisfy the algebraic independence between transformations, $F_0$ and $F_1$, when the corresponding transformations projected in the latent space, $f_0$ and $f_1$, satisfy the algebraic independence.

\paragraph{Theorem 1 (projection of transformations)} \quad \\
Given $F_0 (\boldsymbol{\lambda}_0 )=G_N f_0 (\boldsymbol{\lambda}_0 ) G_P$ and $F_1 (\boldsymbol{\lambda}_1 )=G_N f_1 (\boldsymbol{\lambda}_1 ) G_P$, where $G_P$ is an encoder and $G_N$ is an bijective decoder, if transformations $f_0$ and $f_1$, in the latent spaces satisfy the algebraic independence and the following $G_P-F$ commutativity holds, then the corresponding transformations, $F_0$ and $F_1$, also satisfy the algebraic independence.

$G_P-F$ commutativity: 
\begin{equation}
G_P F_0 (\boldsymbol{\lambda}_0 )[X]=G_P G_N f_0 (\boldsymbol{\lambda}_0 ) G_P [X]=f_0 (\boldsymbol{\lambda}_0 ) G_P [X]
\label{eq:gpf1}
\end{equation}
\begin{equation}
G_P F_1 (\boldsymbol{\lambda}_1 )[X]=G_P G_N f_1 (\boldsymbol{\lambda}_1 ) G_P [X]=f_1 (\boldsymbol{\lambda}_1 ) G_P [X]
\label{eq:gpf2}
\end{equation}

\paragraph{Proof} \quad\\
Commutativity is obtained from the $G_P-F$ commutativity and the above definitions as follows:
\begin{gather}
F_0 (\boldsymbol{\lambda}_0 ) F_1 (\boldsymbol{\lambda}_1 )[X] \notag \\
=G_N f_0 (\boldsymbol{\lambda}_0 ) G_P G_N f_1 (\boldsymbol{\lambda}_1 ) G_P [X] \notag \\
=G_N f_0 (\boldsymbol{\lambda}_0 ) f_1 (\boldsymbol{\lambda}_1 ) G_P [X] \notag \\
=G_N f_1 (\boldsymbol{\lambda}_1 ) f_0 (\boldsymbol{\lambda}_0 ) G_P [X] \notag \\
=G_N f_1 (\boldsymbol{\lambda}_1 ) G_P G_N f_0 (\boldsymbol{\lambda}_0 ) G_P [X] \notag \\
=F_1 (\boldsymbol{\lambda}_1 ) F_0 (\boldsymbol{\lambda}_0 )[X]
\end{gather}

Similarly, the existence of a unit element is obtained, because $f_0$ and $f_1$ satisfy the condition 1 for algebraic independence:
\begin{gather}
F_0 (0) F_1 (\boldsymbol{\lambda}_1 )[X]  \notag \\
=G_N f_0 (0) G_P G_N f_1 (\boldsymbol{\lambda}_1 ) G_P [X] \notag \\
=G_N G_P G_N f_1 (\boldsymbol{\lambda}_1 ) G_P [X]  \notag \\
=G_N f_1 (\boldsymbol{\lambda}_1 ) G_P [X] \notag \\
=F_1 (\boldsymbol{\lambda}_1 )[X]
\end{gather}

\begin{gather}
F_1 (0) F_0 (\boldsymbol{\lambda}_0 )[X] \notag \\
=G_N f_1 (0) G_P G_N f_0 (\boldsymbol{\lambda}_0 ) G_P [X] \notag \\
=G_N G_P G_N f_0 (\boldsymbol{\lambda}_0 ) G_P [X] \notag \\
=G_N f_0 (\boldsymbol{\lambda}_0 ) G_P X \notag \\
=F_0 (\boldsymbol{\lambda}_0 )[X]    
\end{gather}

The uniqueness of the transformation parameters is obtained from the $G_P-F$ commutativity and the above definitions, as follows: 
If we defined the bijective function, $\phi$, as $\phi (\boldsymbol{\lambda}_0,\boldsymbol{\lambda}_1,X)=f_0 (\boldsymbol{\lambda}_0 ) f_1 (\boldsymbol{\lambda}_1 ) G_P [X]$, then
\begin{gather}
\Phi (\boldsymbol{\lambda}_0,\boldsymbol{\lambda}_1,X) \notag \\
= F_0 (\boldsymbol{\lambda}_0 ) F_1 (\boldsymbol{\lambda}_1 )[X] \notag \\
= G_N f_0 (\boldsymbol{\lambda}_0 ) G_P G_N f_1 (\boldsymbol{\lambda}_1 ) G_P [X] \notag \\
=G_N f_0 (\boldsymbol{\lambda}_0 ) f_1 (\boldsymbol{\lambda}_1 ) G_P [X] \notag \\
=G_N \phi(\boldsymbol{\lambda}_0,\boldsymbol{\lambda}_1,X) ,
\end{gather}
where $\Phi$ is the bijective function, because both $G_N$ and $\phi$ are bijective. 
In this way, we obtained the desired result.

\paragraph{Design of latent space transformation} \quad\\
Next, we designed the transformations in the latent space that would allow the algebraic independence to hold. We defined the latent vectors, $\mathbf{x}$ and $\mathbf{y}$, on $\mathbb{R}^{(n_0+n_1 )}$ as follows:

\begin{gather}
\mathbf{x}=(\mathbf{x}_0,\mathbf{x}_1 )=(G_{P0} [X],G_{P1} [X])=G_P [X] \\
\mathbf{y}=(\mathbf{y}_0,\mathbf{y}_1 )=(G_{P0} [Y],G_{P1} [Y])=G_P [Y] \\
\mathbf{x}_0,\mathbf{y}_0 \in Q_0 \\
\mathbf{x}_1,\mathbf{y}_1 \in Q_1 
\end{gather}

Consider the transformations $f_0 (\boldsymbol{\lambda}_0):\mathbf{x}_0  \to \mathbf{y}_0$ and $f_1 (\boldsymbol{\lambda}_1):\mathbf{x}_1 \to \mathbf{y}_1$, which are transformations on different latent spaces. These transformations satisfy the algebraic independence. Then, the $G_P-F$ commutativity (equation \eqref{eq:gpf1} and equation \eqref{eq:gpf2}) can be rewritten as 
\begin{gather}
(\mathbf{y}_0^{'},\mathbf{x}_1^{'} ) \stackrel{\mathrm{def}}{=} G_P G_N (\mathbf{y}_0,\mathbf{x}_1 ) \notag \\
=(\mathbf{y}_0,\mathbf{x}_1 ) 
\end{gather}
and 
\begin{gather}
(\mathbf{x}_0^{'},\mathbf{y}_1^{'} ) \stackrel{\mathrm{def}}{=} G_P G_N (\mathbf{x}_0,\mathbf{y}_1 ) \notag \\
=(\mathbf{x}_0,\mathbf{y}_1 ),
\end{gather}, 
because $f_0 (\boldsymbol{\lambda}_0 ) G_P [X]=(\mathbf{y}_0,\mathbf{x}_1)$ and $f_1 (\boldsymbol{\lambda}_1 ) G_P [X]=(\mathbf{x}_0,\mathbf{y}_1)$ by definitions.

\begin{figure}[t]
\centering
\includegraphics[width=0.9\columnwidth]{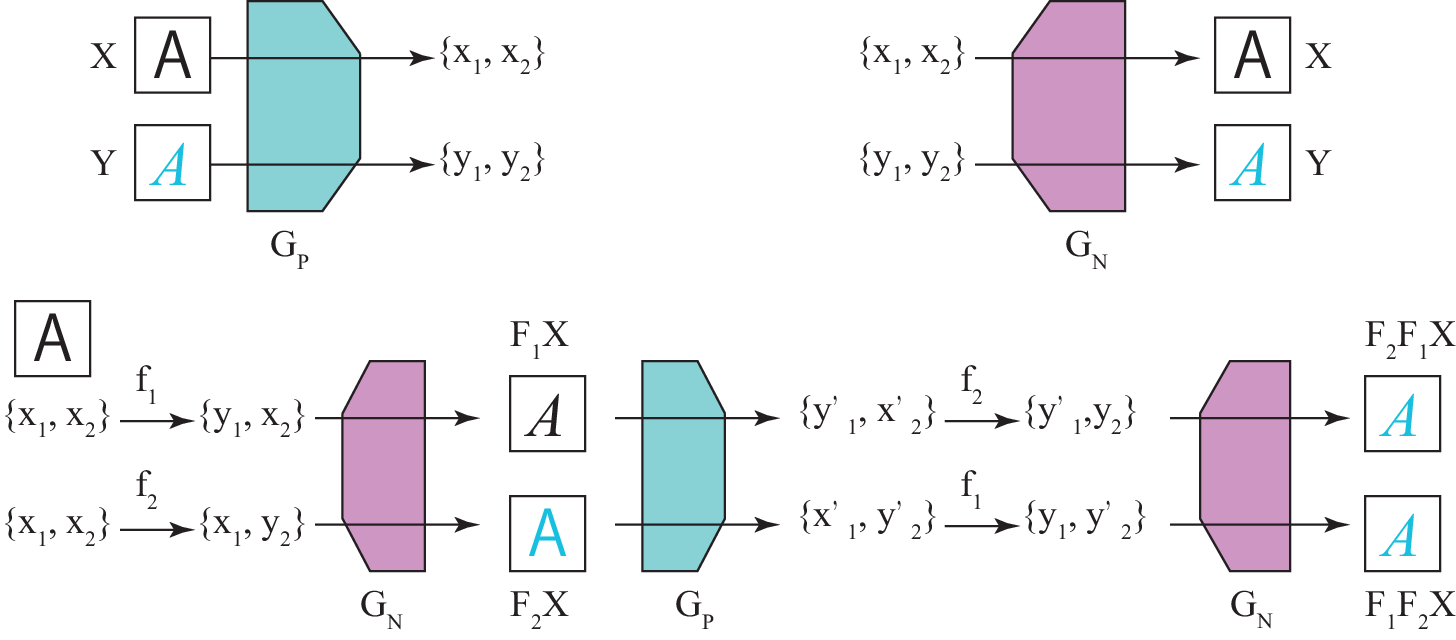}
\caption{Information flow. The network consists of two networks, $G_P$ and $G_N$, where $G_P$ is an encoder from the sensory input to the latent space, and $G_N$ is a decoder from the latent space to the sensory input. Transformations $f_0$ and $f_1$ in the latent space transform latent vector $x$ into latent vector $y$.}\label{fig1}
\end{figure}

To reconstruct the observation $Y$ from the latent vectors, $Y=G_N (\mathbf{y}_0,\mathbf{y}_1)$ needs to be maintained. Combining the reconstruction condition and $G_P-F$ commutativity, we obtain the following loss functions, 
$\| Y-G_N (\mathbf{y}_0,\mathbf{y}_1^{'})\|$ and $\| Y-G_N (\mathbf{y}_0^{'},\mathbf{y}_1)\|$. 
Figure \ref{fig1} shows a summary of our encoder–decoder model. 
If $Y=G_N (\mathbf{y}_0,\mathbf{y}_1^{'} )=G_N (\mathbf{y}_0^{'},\mathbf{y}_1 )$, then $\mathbf{y}_0=\mathbf{y}_0^{'}$ and $\mathbf{y}_1=\mathbf{y}_1^{'}$ in the $G_P-F$ commutativity hold, because $G_N$ is the bijective function. 
The same is true for $X$.
After $G_P$ and $G_N$ learning to satisfy the loss function, we obtained two distances between $X$ and $Y$ from the projections as follows:
\begin{gather}
\| Y-X \|_0 \stackrel{\mathrm{def}}{=} \|  G_{P0} Y-G_{P0} X\| = \| \mathbf{y}_0-\mathbf{x}_0 \| \\
\| Y-X \|_1 \stackrel{\mathrm{def}}{=} \| G_{P1} Y-G_{P1} X\| =\| \mathbf{y}_1-\mathbf{x}_1 \|
\end{gather}
In this theory, both transformations $f_0$ and $f_1$ in the latent spaces are maps between two latent vectors projected from the countable set, $S$. Therefore, we do not assume the transformations are neither continuous groups nor the Lie groups, which are often used in invariant transformation theories\cite{cohen2014, takada2021}.

\section{Method}
\paragraph{Dataset} \quad \\
The dataset consists of 26 alphabets with 12 fonts and 7 colors. The image size is 3 channels $\times$ 32 pixels $\times$ 32 pixels. The background color is black, (0,0,0) in (R, G, B).  The seven colors of alphabets employed consist of (0, 0, 1), (0, 1, 0),$\ldots$, (1, 1, 1). Two images are randomly sampled to make pairs of $(X, Y)$. To increase the variety of colors, a random value from 0.2 to 1 was multiplied to the color channels. 

\paragraph{Network Structure} 
\subparagraph{$G_{P0}, G_{P1}:$} 
The input image is convolved using three Convolutional Neural Networks (CNN)\cite{krizhevsky2012} whose kernel size is 4, stride is 2 and padding is 1 without bias, and two linear layers follow. We used ReLU function for activation. The channels of the convolution layers were 128, 256 and 512. The output dimensions of the linear layers were 4096 and 32. The number of dimension of the latent vectors was 32. 

\subparagraph{$G_N:$} 
To realize the bijective function, we employed the globally injective ReLU Network\cite{Puthawala2022}. The input was a tuple of two latent vectors ($\mathbf{x}_0$,$\mathbf{x}_1$). The network consisted successively of three linear layers without bias and three transposed convolution layers whose kernel size is 4, stride is 2, and padding is 1 without bias, and final layer is Conv1$\times$1 layer\cite{lin2014}. The output channels of the linear layers were 128, 1024 and 4096. 
In the transposed convolution layer, when convolution matrix is denoted by $C$ whose input is  ($ch_{in}$,$width$,$height$)-size tensor and output is ($ch_{in}/4$,$width*2$,$height*2$)-size tensor, the layer is defined by $(C,-C)$. We set ch to 32. We used the same network configuration using different initializations for comparison.

\paragraph{Loss functions} \quad \\
Let points $X$ and $Y$ be on observation space $S \in \mathbb{R}^N$. The latent vectors are calculated by two independent encoders, $G_{P0}$ and $G_{P1}$, as follows: 

\begin{gather}
(\mathbf{x}_0,\mathbf{x}_1)=(G_{P0}X,G_{P1}X), \\
(\mathbf{y}_0,\mathbf{y}_1)=(G_{P0}Y,G_{P1}Y).
\end{gather}

Using the single-transformation images defined by the equations \eqref{st-image}\eqref{st-image2}, 
\begin{gather}
\label{st-image}
F_{0} X=G_{N} (\mathbf{y}_0,\mathbf{x}_1) \\
\label{st-image2}
F_{1} X=G_{N} (\mathbf{x}_0,\mathbf{y}_1), 
\end{gather}
we calculated the latent vectors of the constructed single-transformation images as follows: 
\begin{gather}
\mathbf{y}_{0}^{'}=G_{P0} F_{0} X,\\
\mathbf{y}_{1}^{'}=G_{P1} F_{1} X.
\end{gather}
Finally, we calculated the reconstruction images as follows: 
\begin{gather}
F_{1} F_{0} X=G_{N} (\mathbf{y}_{0},\mathbf{y}_{1}^{'} ), \\ 
F_{0} F_{1} X=G_{N} (\mathbf{y}_{0}^{'},\mathbf{y}_{1}). 
\end{gather}
We defined the loss function as loss=$\| Y-F_{1} F_{0} X\| +\alpha \|Y-F_{0} F_{1} X\| $, where $\alpha$ is hyper parameter. We set $\alpha$ to 1 in the typical condition and to 0 in the ablation condition.

\paragraph{Training} \quad \\
We used RAdam\cite{liu2020} for optimization. 
The learning rate was 1e-4 and batch size was 128. We used CUDA 11.4, PyTorch 1.10.0 \cite{paszke2019} and Nvidia RTX 
3080Ti for training. 
Training epochs were 400 for comparison experiments or 1000 for other experiments. 

\paragraph{Evaluation} \quad \\
We evaluated the single-transformation images that are constructed from the input $X$ and $Y$ by the following invariance index. If $F_0$ and $F_1$ are successfully learned, each transformation becomes a color-only transformation or a shape-only transformation.  The color-only transformation must keep the shape invariant. And the shape-only transformation must keep the color invariant. 

\subparagraph{Color invariance: } 
Each channel in the single-transformation images, $F_0 X$ and $F_1 X$, was binarized by threshold $\tau_c$. We set $\tau_c$=0.1. Consequently, the binarized images had eight colors (0, 0, 0), (0, 0, 1),$\ldots$, (1, 1, 1). We defined the most frequent color, except background color (0, 0, 0), as the letter color. We calculated the inner product of normalized vectors between the letter color of $X$ and that of $F_0 X$ or $F_1 X$. We defined the mean of the results in the batch as color invariance. The color invariance is close to 1 when the transformation result does not change the color of the input image. We set batch size to 128. 

\subparagraph{Shape invariance: }
The single-transformation images, $F_0 X$ and $F_1 X$, were binarised by threshold $\tau_s$. We set $\tau_s$=0.1. We defined the mean of the inner product of normalized vectors in the batch between the binarized image of input $X$ and the binarized image of $F_0 X$ or $F_1 X$ as the shape invariance. The shape invariance is close to 1 when the transformation result does not change the shape of input image. 

\subparagraph{Invariance: }
We defined a transformation with a lower color invariance than another transformation as the shape transformation. Further, we defined a transformation with a lower shape invariance than another transformation as the color transformation. We defined the invariance as the mean of a color invariance of the shape transformation and a shape invariance of the color transformation. 

\section{Results}
We verified that the neural network system which would learn the algebraic independence structure between neural networks could represent the invariant transformation between sensory information and construct multiple metric spaces. We assumed $G_{P0}$, $G_{P1}$ and $G_{N}$ in the above formulation were neural networks with network parameters trained to satisfy the $G_P-F$ commutativity. Therefore, we employed the globally injective ReLU\cite{Puthawala2022} network for the bijective function, $G_N$. To satisfy the $G_P-F$ commutativity, we employed the following loss function: 
\begin{equation}
\| Y-G_{N} (\mathbf{y}_0,\mathbf{y}_{1}^{'} )\|+\| Y-G_{N} (\mathbf{y}_{0}^{'},\mathbf{y}_{1})\|,
\end{equation}

where 
\begin{gather}
F_0 F_1 X \stackrel{\mathrm{def}}{=} G_N (\mathbf{y}_{0},\mathbf{y}_{1}^{'} )=G_N (G_{P0} Y,G_{P1} G_{N} (G_{P0} X,G_{P1} Y)), \\ 
F_1 F_0 X \stackrel{\mathrm{def}}{=} G_N (\mathbf{y}_{0}^{'},\mathbf{y}_{1} )=G_N (G_{P0}
 G_{N} (G_{P0} Y,G_{P1} X),G_{P1} Y).
\end{gather}

Our training dataset consisted of 26 alphabets with 12 fonts and 7 colors. Data X and Y were randomly sampled from this dataset. To increase the variety of colors, a random value was multiplied by the color channels of the sampled data. We obtained two single-transformation images, $F_0 X$ and $F_1 X$, which were defined as follows: 
\begin{gather}
F_0 X=G_N (G_{P0} Y,G_{P1} X), \\
F_1 X=G_N (G_{P0} X,G_{P1} Y).
\end{gather}
These images are the result of a single transformation applied to the input. These are expected to be single feature-type transformations resulting from learning. 

\begin{figure}[t]
\centering
\includegraphics[width=0.9\columnwidth]{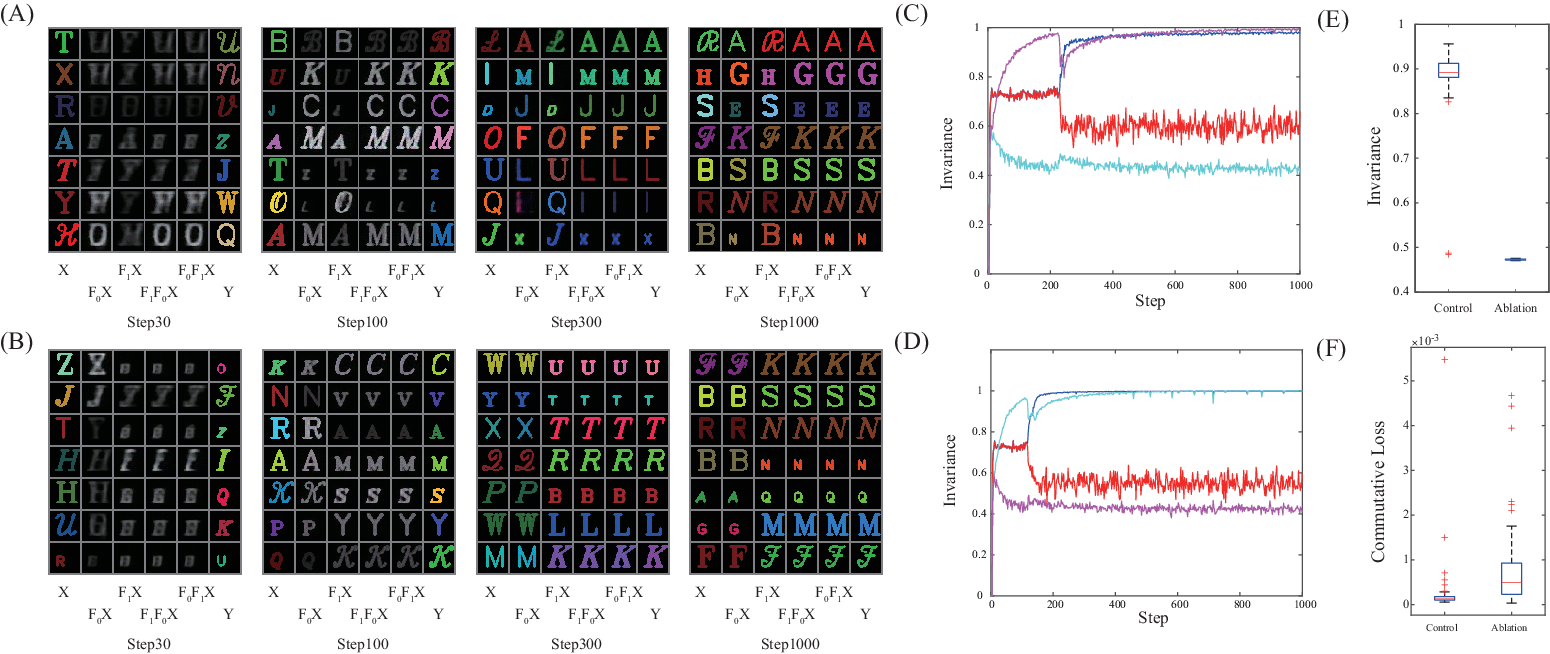}
\caption{A brief timeline of the algebraic independence learning. (a) Results with commutative learning. $F_0 X$ converged to the shape transformation. $F_1 X$ converged to the color transformation. (b) A typical example of an ablation study. Both color and shape transformations were conducted by only $F_1 X$. (c) Learning changes in invariance in the control condition. (d) Same changes in the ablation condition. Blue, color invariance of $F_0 X$; red, color invariance of $F_1 X$; cyan, shape invariance of $F_0 X$; and magenta, shape invariance of $F_1 X$. (e) Median invariances in the control and ablation groups were 0.8919 and 0.473; the distributions in the two groups differed significantly (Mann–Whitney U = 10000, n = 100, P $<$ $2.5 \times 10^{-34}$, two-tailed). (f) Median commutative loss after training in the control and ablation groups were 1.275e-4 and 4.934e-4; the distributions in the two groups also differed significantly (Mann–Whitney U = 1869, n = 100, P $<$ $2.025 \times 10^{-4}$, two-tailed).}\label{fig2}
\end{figure}

In the control condition, two images, $F_0 X$ and $F_1 X$, became the results of either a color-only transformation or a shape-only transformation through training. 
We observed that the color-only transformation changes only hue and not the contrast (Fig. \ref{fig2}a). In two of 100 trials, single feature transformations could not be learned, and a both types transformation and an identity function were learned because the identity function satisfies the condition of algebraic independence between arbitrary transformations. 
In the ablation condition, wherein only a reconstruction loss, $\| Y-F_1 F_0 X\|$, is learned, single-transformation images, $F_0 X$, $F_1 X$ could not become single feature transformations. In all trials, a both types transformation and an identity function were learned (Fig. \ref{fig2}b). $F_0$ became the identity function, and $F_0 X$ became equal to $X$. $F_1$ became the both types transformation, and $F_1 X$ became equal to $Y$.
We evaluated the color and shape invariances of the single transformation images (Fig. \ref{fig2}c-d). The invariance during training reflects the timeline of single transformation learning (Fig. \ref{fig2}a-b). The invariance, which corresponds to the mean of the color invariance of the shape transformation and the shape invariance of the color transformation, significantly differed between the control condition and the ablation condition after training (Fig. \ref{fig2}e, Mann–Whitney U = 1000, n = 100, P $< 2.5\times 10^{-34}$, two-tailed). The outliers in Fig.\ref{fig2}e reflects two failure cases. Similarly, the $G_P-F$ commutativity loss after training significantly differed between conditions (Fig. \ref{fig2}f, Mann–Whitney U = 1869, n = 100, P $< 2 \times 10^{-14}$, two-tailed). Thus, the $G_P-F$ commutativity learning is required to stably categorize the color and shape transformation.

\begin{figure}[t]
\centering
\includegraphics[width=0.7\columnwidth]{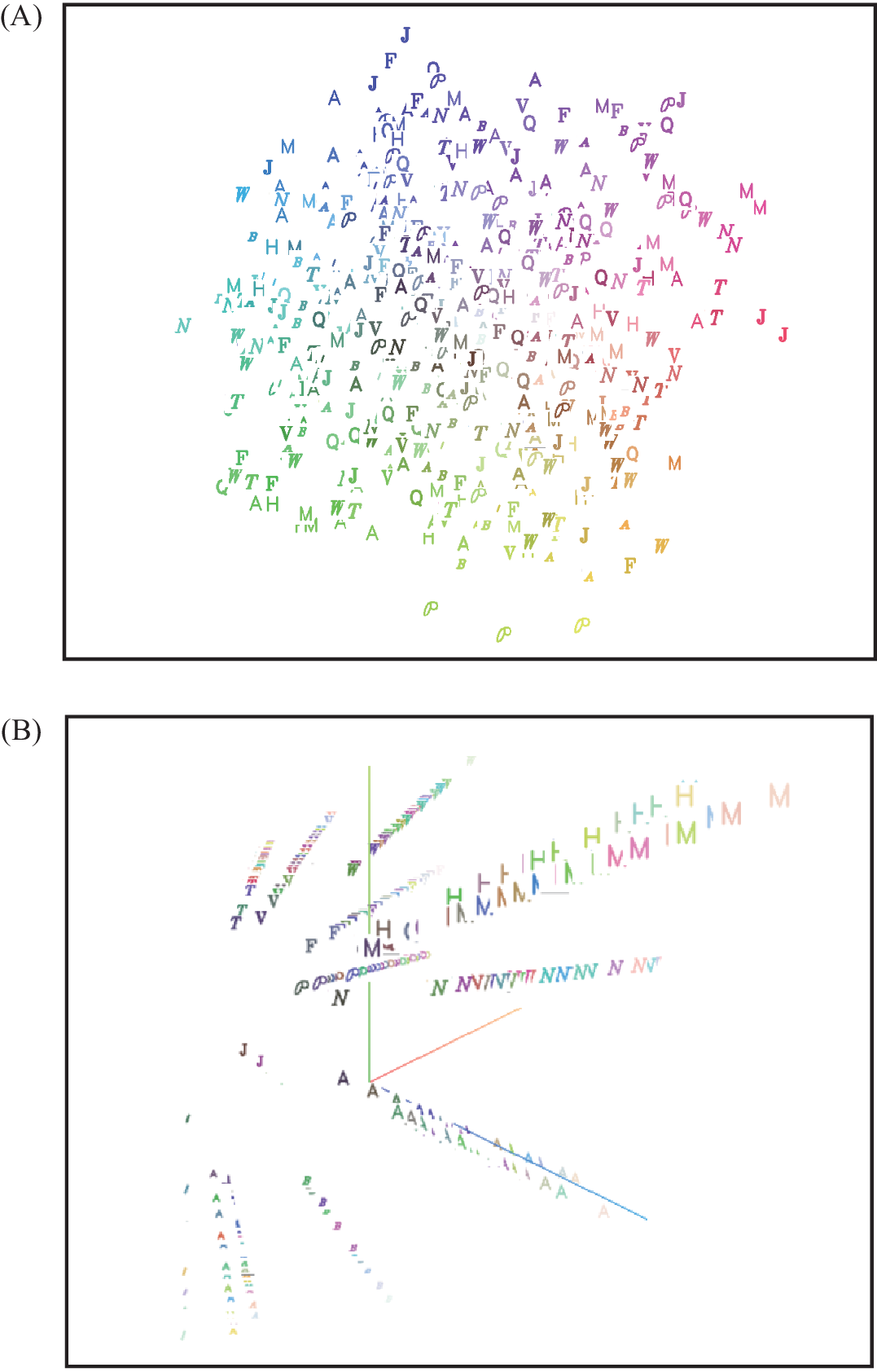}
\caption{ Learned metric spaces. (a) Metric space corresponding to the color transformation. (b) Metric space corresponding to the shape transformation. Sixteen different letters in 32 different colors were plotted to the learned metric spaces. Each space was mapped in 2D using PCA for visualization. Contribution rate: 99.3\% in the color space, 64.8\% in the shape space. For visualization, the background color was changed from black to white.}\label{fig3}
\end{figure}

After training, two 32 dimension latent vectors corresponding to the transformations were projected to two or three dimensional spaces using principle component analysis for visualization (Fig. \ref{fig3}). In color feature space (Fig. \ref{fig3}a), the same color with different shapes was arranged into clusters. In shape feature space (Fig. \ref{fig3}b), the same shapes with different color were arranged into clusters. Furthermore, we observed that the same shape data were linearly arranged from the origin in the shape feature space. Thus, the latent vectors could be divided into two different metric spaces corresponding to color and shape.

\section{Discussion}
In this study, we established the connection between the algebraic independence of transformations and the invariant transformation equation. From our point of view, the transform parameter $\lambda$ was limited to scalars with conventional axis independence. With algebraic independence, the transformation parameter is generalized from scalars to vectors. Subsequently, we tested whether the artificial neural network system can autonomously acquire the invariant transformation equation through representation learning to satisfy the algebraic independence between neural network modules and project sensory information onto multiple metric spaces to independently evaluate the differences and similarities in each feature. 
Our theory provides general requirements for the unsupervised categorization of independent similarity measures corresponding to sensory features. 
In our experiments, all ablation conditions failed to learn a single-feature transformation, indicating that color and shape spaces cannot be separated. In contrast, most of the control conditions were able to learn a single-feature transformation and can separate the metric spaces corresponding to color and shape. However, a few cases in the control condition failed to learn and a transformation became the identity function, which also satisfies the algebraic independence between arbitrary transformations.
To avoid learning the identity transformation, it is believed that further constraints are needed. Solving this problem is an important topic for future research.
Notably, our experiment is limited to the categorization of two features, and our formulation assumes that the sensory input has only a single object in a scene. When the number of transformations is greater than two, algebraic independence can be defined between arbitrarily chosen pairs of transformations. With such a complex algebraic structure, extensive research is required to construct a more realistic representation structure.  Regardless of such limitations, our mathematical theory of categorizing similarity measures shows how the neural network system can autonomously learn the permanence of perception through algebraic structure learning between neural networks. This suggests that our theory can contribute to the further development of the mathematical theory of functional differentiation of neural networks. 





\bibliographystyle{elsarticle-num} 
\bibliography{ AlgebraicIndependence.bib }






\section*{Code availability}
Source codes are available at \\
https://github.com/Yoshiyuki-Ohmura/CommutativeLearning.

\section*{Acknowledgements}
This paper is based on results obtained under a Grant-in-Aid for Scientific Research (A) JP22H00528. 

\section*{Author Contributions}
YO substantially contributed to the study conceptualization and the mathematical theory. WS and YO contributed to the experimental design. WS developed a part of software, neural networks design and dataset. YO mainly developed the software, neural networks design and loss function. YO wrote original draft preparation. YK supervised the project. All authors discussed the results and commented on the manuscript. 

\section*{Competing Interests Statement}
The authors declare no conflicts of interest associated with the manuscript. 

\end{document}